\documentclass[letterpaper]{article} 
\usepackage{aaai24}  
\usepackage{times}  
\usepackage{helvet}  
\usepackage{courier}  
\usepackage[hyphens]{url}  
\usepackage{graphicx} 
\usepackage{natbib}  
\usepackage{caption} 
\usepackage{algorithm}
\usepackage{algorithmic}
\usepackage{xcolor}
\usepackage{multirow}
\usepackage{subfig}
\usepackage{amsmath}
\usepackage{amssymb}
\usepackage{booktabs}

\usepackage{newfloat}
\usepackage{listings}
\DeclareCaptionStyle{ruled}{labelfont=normalfont,labelsep=colon,strut=off} 
\floatstyle{ruled}
\newfloat{listing}{tb}{lst}{}
\floatname{listing}{Listing}
\nocopyright

\title{ViCo: Engaging Video Comment Generation with Human Preference Rewards}
\author {
    Yuchong Sun\textsuperscript{\rm 1},
    Bei Liu\textsuperscript{\rm 2}, 
    Xu Chen\textsuperscript{\rm 1}, 
    Ruihua Song\textsuperscript{\rm 1}\thanks{Corresponding author}, 
    Jianlong Fu\textsuperscript{\rm 2}, 
}
\affiliations {
    \textsuperscript{\rm 1} Renmin University of China\\
    \textsuperscript{\rm 2} Microsoft Research Asia\\
    \{ycsun, xu.chen, rsong\}@ruc.edu.cn, \{bei.liu, jianf\}@microsoft.com
}

\usepackage{bibentry}

\begin{document}

\maketitle

\begin{abstract}
Engaging video comments
play an important role in video social media, as they are the carrier of feelings, thoughts, or humor of the audience. 
Preliminary works have made initial exploration for video comment generation by adopting caption-style encoder-decoder models.
However, comment generation presents some challenges different from caption generation, which makes these methods somewhat less effective at generating engaging comments.
In contrast to the objective and descriptive nature of captions, comments tend to be inherently subjective, making it hard to quantify and evaluate the engagement of comments. 
Furthermore, the scarcity of truly engaging comments brings difficulty to collecting enough high-quality training examples.
In this paper, we propose \textbf{ViCo} with three novel designs to tackle the above challenges for generating engaging \textbf{Vi}deo \textbf{Co}mments.
Firstly, to quantify the engagement of comments, we utilize the number of ``likes'' each comment receives as a proxy of human preference after an appropriate debiasing procedure.
Secondly, to automatically evaluate the engagement of comments, we train a reward model to align its judgment to the above proxy. Our user studies indicate that this reward model effectively aligns with human judgments.
Lastly, to alleviate the scarcity of high-quality comments, an initial generator is trained on readily available but noisy data to generate comments. Then the reward model is employed to offer feedback on the generated comments, thus optimizing the initial generator.
To facilitate the research of video commenting, we collect a large video comment-dataset (ViCo-20k) with rich metadata from a popular video website.
Experiments on ViCo-20k show that the comments generated by our ViCo model exhibit the best performance in terms of both quantitative and qualitative results, particularly when engagement is considered.

\end{abstract}
\section{Introduction}
When we watch videos on the Internet, such as YouTube and TikTok, we are often drawn to comments underneath the videos that highlight emotionally resonant moments, express opinions in a humorous way, or raise engaging and thought-provoking topics.
Automatically creating such engaging comments for a video is a novel and essentially critical research problem. It not only tests the capabilities of artificial intelligence in multimodal understanding and generation, but also has far-reaching benefits for downstream applications, such as grounded multi-modal dialogues, affective computing, and the metaverse.

Recent advanced multi-modal models~\cite{lin2022swinbert,seo2022endcaption,tang2021clip4caption, wang2022git} are good at describing the content of videos. They can describe the persons, objects, and events unfolding within the video. 
\begin{figure}
    \centering
    \includegraphics[width=\columnwidth]{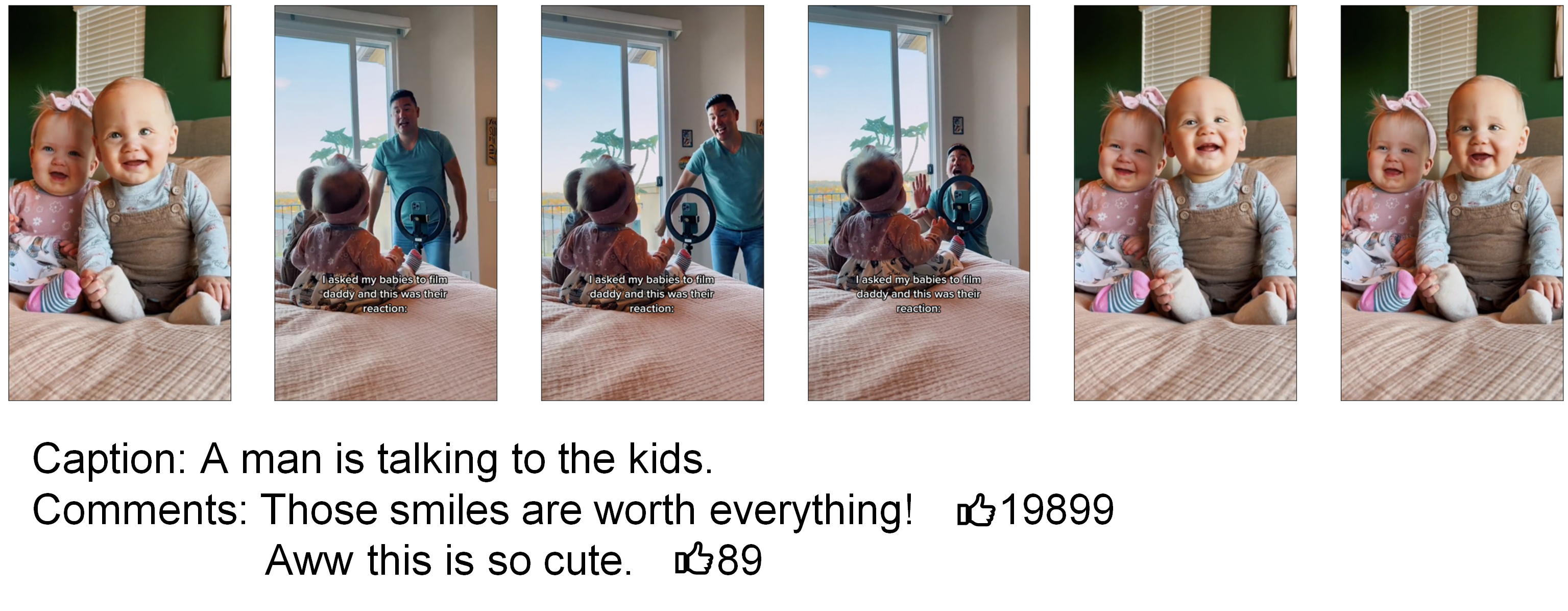}
    \caption{An example of video caption and comment. Video captioning aims to describe the concrete events in the video, but an engaging comment describes the most touching thing of the video (e.g., the smile of the kids in the video). The other generic comment is however liked by fewer people.}
    \label{figs:task_case}
\end{figure}
Nevertheless, captions aim to reflect basic perception ability and describe the things that are often obvious and objective, which may not engage audiences~\cite{shuster2019engaging}. As illustrated in Fig.~\ref{figs:task_case}, while the video caption offers a description of the concrete events in the video, it is worth noting that a human-preferred comment, mentioning the most touching thing in the video to be the children's smiles, received significantly more engagement as compared with the general comments, such as simply describing the children as cute.

Although generating comments and captions both involve generating text for videos, generating video comments raises distinct challenges compared to video captioning, especially when requiring them to engage people.
Different from video captioning which maps visual events to objective and descriptive text, comments are subjective and diverse. This makes it hard to quantify and evaluate the engagement of comments.
Furthermore, there is a serious long-tail phenomenon in the distribution of comments. Although there are many video comments, only a small portion is considered engaging, and even worse, most of them are available for only a small portion of videos.
Some recent works~\cite{ma2019livebot,wang2020videoic,marrese2022open-commentary} have explored using caption-style encoder-decoder models for video comment generation.
Due to the fact that these captioning-style models merely mimic the distribution of comments without considering their quality, resulting in the generation of too common comments (e.g., ``so cute'', and ``wow'') that are prevalent in the training corpus.

To tackle the above challenges in generating engaging video comments, we propose a model called ViCo with three novel designs.
First, to quantify the engagement of a video comment, we propose to use the number of ``likes'' as a proxy measurement with a debiased processing step. They collect real-user feedback and naturally reflect the opinions of the crowd.
Second, to automatically evaluate the engagement of video comments, we introduce a reward model to judge whether the comments are engaging or not. To make the reward model consistent with human preference, we train it based on the real-user feedback in a pairwise-comparison manner. Our user studies demonstrate this reward model is well aligned with real-human judgment.
Third, to alleviate the scarcity of high-quality comments, we use an initial generator trained on readily
available but noisy data to generate comments and use the reward model to assess the quality of the generated comments, and then optimize the generator based on the reward feedback.
To facilitate future research, we collect a large video-comment dataset (ViCo-20k) with rich metadata from a popular video website\footnote{https://www.tiktok.com}.  
We conduct thorough experiments over the dataset to verify the effectiveness of our methods. The results indicate that our proposed method improves the richness of generated comments, and the reward model can guide the generator to produce comments that are favored by humans over baseline methods.

Our contributions are summarized as follows: 
\begin{itemize}
    \item We propose to adopt the number of likes in each comment as a proxy to quantify the engagement of video comments.
    \item We train a reward model to align its judgments of comment engagement with human preference, enabling automated evaluation of video comments.
    \item We utilize the reward model to assess the generated comments of an initial generator, and then optimize the generator based on the reward feedback.
    \item We collect a large video-comment dataset (ViCo-20k) with rich metadata. It will be public to facilitate the research community.
\end{itemize}

\section{Related Works}

\subsection{Video Caption Generation}
Video captioning~\cite{xu2016msrvtt,krishna2017actnetcaption} aims to describe the visual contents and events in videos. Earlier methods adopt an encoder-decoder architecture, where a CNN is used to extract vision features and an RNN is used to predict the words. 
Inspired by the success of pre-training paradigm in NLP~\cite{Devlin2018bert,radford2019gpt2,brown2020gpt3}, recent methods adopt the transformer-based architecture, and pre-training on large-scale vision-language data for caption generation and achieve leading performance~\cite{wang2022git,tang2021clip4caption}.
Recently, some works combine large vision models (LVMs) and large language models and achieve strong multimodal reasoning and generation abilities~\cite{alayrac2022flamingo,tsimpoukelli2021frozen,mokady2021clipcap,li2023blip-2}. Inspired by their success, we also adopt a LVM to extract high-quality visual features and use the strong reasoning and generation abilities of LLM in the video comment generation task.

\subsection{Video Comment Generation}
Most existing studies~\cite{ma2019livebot,wang2020videoic,marrese2022open-commentary} on video comment generation focus on creating live or time-synchronized comments, known as danmaku or barrage, which are prevalent on Chinese video platforms but less on English ones. 
This task aims to generate new comments for specific moments in a video, using nearby video frames and comments as context. 
Livebot~\cite{ma2019livebot} proposes the first danmaku dataset and Unified Transformer model for this task, which uses a video encoder and a text encoder to process video and comments respectively, and a decoder to generate new comments. 
The following works~\cite{wang2020videoic,zeng2021plvcg} also depend on this encoder-decoder architecture but design methods for enhancing interaction between the input modalities.
Different from time-synchronized comments or live comments in LiveBot~\cite{ma2019livebot} and VideoIC~\cite{wang2020videoic}, we focus on a more common form of comments where we neither know which frames a comment is evoked from, nor do we have relevant comments issued by other users, and thus it is more challenging. 
Furthermore, our methods have been specifically designed to generate engaging comments.

\subsection{Reinforcement Learning for Text Generation}
Reinforcement learning has been adopted to solve the misalignment of training loss and evaluation metrics in text generation tasks~\cite{rennie2017scst,ranzato2015sequence}. In machine translation~\cite{ranzato2015sequence} and image caption tasks~\cite{rennie2017scst}, automatic metrics such as BLEU~\cite{papineni2002bleu} and CIDEr~\cite{vedantam2015cider} are widely used to calculate rewards which are used to optimize the generator.
Recently, Reinforcement Learning from Human Feedback (RLHF) achieves great success in NLP areas, such as ChatGPT~\cite{openai2022chatgpt} and GPT-4~\cite{OpenAI2023GPT4TR}, which use human-annotated ranking data to train a reward model and then using the reward model to compute rewards during the reinforcement learning stage. RLHF can make the model better understand human instructions and generate better responses. However, collecting human-annotated rankings can be extremely costly. In this paper, we are the first to use RLHF to improve the engagement of generated video comments, and instead of using manual annotation, we use the number of likes as a proxy of human preference for reward model training.
\begin{figure*}[t]
    \centering
    \includegraphics[width=0.95\linewidth]{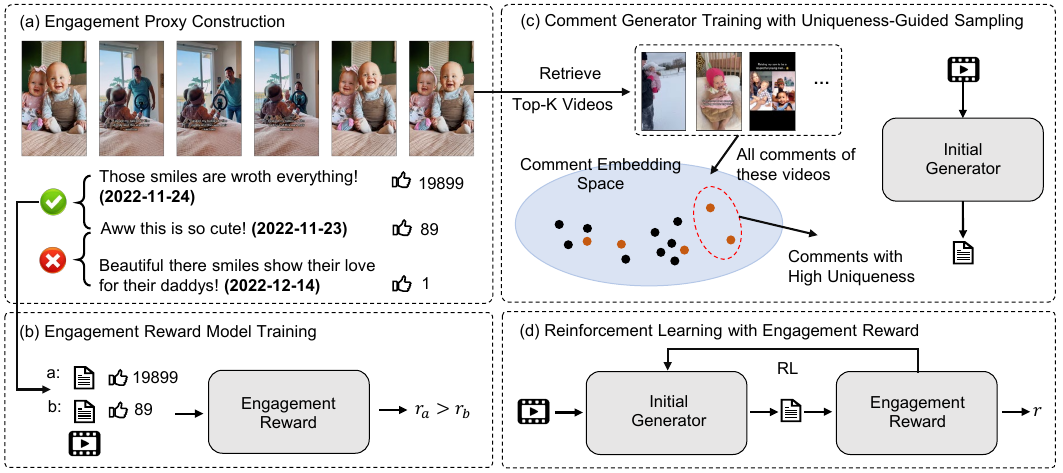}
    \caption{Overall framework of our approaches for generating engaging video comments. (a) We remove the temporal bias of the number of likes and then construct a proxy for the engagement of comments. (b) We use debiased comparison pairs to train an engagement reward model to evaluate the engagement of comments. (c) We train an initial comment generator with uniqueness-guided sampling, which samples more unique comments during training. (d) We use the engagement reward model to guide the initial generator through reinforcement learning.}
    \label{figs:framework}
\end{figure*}

\section{Approach}

\subsection{Constructing a Proxy of Engagement}

Due to the subjectiveness of comments, it is hard to quantify what makes a comment engaging, thus we try to find an automatic proxy of such engagement.
As shown in Fig.\ref{figs:framework} (a), each comment of online videos can be liked by many users, and these likes can be considered the consensus of many audiences about the engagement of the comments. The number of likes typically reflects how much people like the comment. This inspires us to use the number of likes as a proxy for the engagement of video comments. 

\begin{figure}
    \centering
    \includegraphics[width=0.95\linewidth]{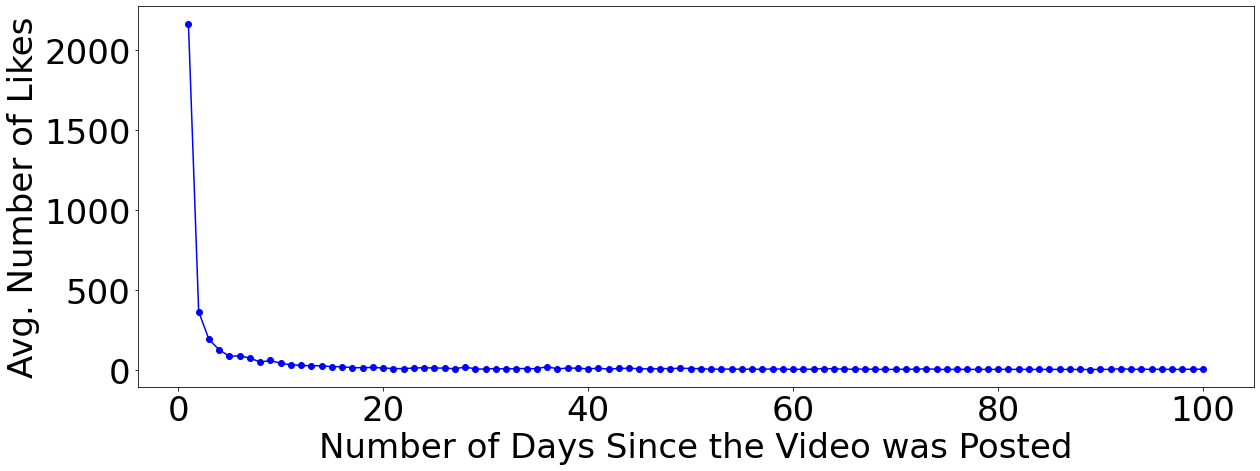}
    \caption{Temporal bias of the original number of likes. The horizontal axis represents the number of days since the video was published, and the vertical axis shows the average number of likes for comments posted on that day.}
    \label{figs:position_bias}
\end{figure}

\textbf{Temporal Debiasing.}
Directly using the number of likes for engagement measurement has a bias problem. It is natural for us to vote for comments appearing at the top. Although there are many factors affecting the ranking algorithm, we find publishing time is one of the key factors. 
By analyzing the number of likes and the publishing time of comments as shown in Fig.\ref{figs:position_bias}, we find that the later the comments are posted, the fewer likes they will receive on average. Therefore, comments with fewer likes may not necessarily be of lower quality. As we show in Fig.~\ref{figs:framework} (a), the third comment has fewer likes, while its quality is significantly higher than the second one.
Therefore, using the raw number of likes as the proxy is unreasonable. 
Based on our observation, the comments posted later but receiving more likes most likely have higher quality. 
Thus we adopt a temporal-aware pair construction method to avoid the temporal bias for the number of likes. 
For two comments in a pair, we ensure the positive comment is published later and has more likes than the negative one.

\subsection{Training Engagement Reward Model}
The engagement of comments is often the result of consensus by many audiences, which makes it difficult to be automatically evaluated.
With the engagement proxy designed above using the number of likes in each comment, we propose to train a reward model to align its measurement to this proxy. The reward model can be further used as an evaluator to judge the engagement of comments.

\textbf{Constructing Comparison Pairs.} 
Different from the recent reinforcement learning from human feedback (RLHF) paradigms in NLP (e.g., ChatGPT~\cite{openai2022chatgpt}, InstructGPT~\cite{ouyang2022instructgpt}), which rely on manual labeling at great cost to collect comparison pairs for training the reward model, we construct many comparison pairs automatically that rely on the debiased number of likes:
\begin{equation}
    D = \{v_i, c_{i,j}, c_{i, j^\prime}\}_{i=1}^{N}, \quad \mathrm{ s.t. } \ e_{i,j}>e_{i,j^\prime}, \ t_{i,j}>t_{i,j^\prime}
\end{equation}
where $D$ is the constructed comparison pairs with $N$ videos, $c_{i,j}$ means $j\mbox{-}th$ comment of video $v_i$, $e_{i,j}$ is the number of likes of $c_{i,j}$, and $t_{i,j}$ means the publish time. We treat $c_{i,j}$ as a positive comment while $c_{i,j\prime}$ as a negative comment.

\textbf{Training the Reward Model.} 
The reward model takes a video and a comment as input and outputs a scalar value to measure the engagement of the comments.
We train the reward model $R_\theta$ using the above comparison pairs to make it assign higher scores to the positive one. The reward model is optimized with pairwise ranking loss as follows:
\begin{equation}
    loss(R_\theta) = -\mathbb{E}[\log(\sigma(r_\theta(v_i, c_{i,j}) - r_\theta(v_i, c_{i,j^\prime})))]
\end{equation}
where $r_\theta(v,c)$ is the scalar output of the reward model for video $v$ and comment $c$.

To verify the effectiveness of the reward model, we sample a number of comparison pairs and ask annotators to compare their engagement. We then calculate the agreement of human judgment and the reward model judgment. The results in Evaluation Section show that the reward model is highly aligned with the judgments of our annotators.

\subsection{Uniqueness-Guided Sampling Method}
Though there are many comments for online videos, a large portion of them may only contain common phrases (e.g., ``cute'', and ``wow''), which will lead the model to generate general and unappealing comments.
To encourage the model to find specific content and generate more unique comments for videos, we design a metric to measure the uniqueness of each comment in the training dataset and then use the uniqueness to guide the sampling of training samples.

\textbf{Uniqueness of Each Comment.}
We assume a video comment that appears in many videos, especially similar types of videos, is more general and unappealing. For example, ``so cute'' appears in many videos about kids, while comments that describe things unique to this video, e.g., the smiles in Fig.~\ref{figs:framework} (a) have higher quality. 
Thus, for each video, we first search the top-$k$ similar videos using the feature of LVM, 
and gather all of the comments belonging to these videos. For each comment, we search the top-$m$ similar comments in the collections and compute the average BERT~\cite{Devlin2018bert} similarity. The lower similarity indicates the comment is more unique to this type of videos.

\textbf{Uniqueness-Guided Sampling.}
The goal of Uniqueness-Guided Sampling is to increase the uniqueness of the sampled comments during generator training. 
For a video $v_i$ and its $J$ comments $c_i=\{c_{i,1}, c_{i,2},..., c_{i,J}\}$, we first sort the comments in descending order of uniqueness.
Then we sample a normalized position index using the following probability density function: 
\begin{equation}
\begin{aligned}
    &P(x) = a + 2(1-a)x, x \in [0,1], \\
    &c_{i,j\prime} = c_{i,\lfloor{x*J}\rfloor}, x\sim P(x),
\end{aligned}
\end{equation}
where $a$ is a hyperparameter to control the probabilities of sampling more small indexes corresponding to more unique comments. As $a$ increases, it tends to sample more unique comments. In particular, when $a$ is set to 1, it degrades to random sampling. $c_{i,j\prime}$ is the sampled comment for training.

\subsection{Reinforcement Learning with Engagement Reward}
After training with the proposed uniqueness-guided sampling comments, the initial generator can generate some comments for videos, while still achieving low engagement.
Because the initial generator is trained to optimize the likelihood of text tokens on noisy data with a small portion of them engaging audiences.

Inspired by the success of the RLHF paradigm in NLP, we combine the initial generator and the reward model in a training loop through reinforcement learning. 
During the training loop, given a video $v_i$, we first sample comments from the outputs of the initial generator:
\begin{equation}
    c_{i,1}, c_{i,2}, ..., c_{i,n} = G_\theta(v_i),
\end{equation}
where $c$ is the sampled comments. Then we use the reward model to measure the engagement of them:
\begin{equation}
    r_{i,1}, r_{i,2}, ..., r_{i,n} = R_\theta(v_i, [c_{i,1}, c_{i,2}, ..., c_{i,n}]),
\end{equation}
where $r$ is a scalar reward for a video-comment pair. Then these $[v_i, c_{i,j}, r_{i,j}]$ can be used to improve the initial generator by optimizing the following objective:
\begin{equation}
    \mathop{max}\limits_{G_\theta} \mathop{\mathbb{E}}\limits_{v\sim V} \left[ \mathop{\mathbb{E}}\limits_{c\sim P(\cdot|v, G_\theta)} R_\theta(v,c)\right],
\end{equation}
where $P(\cdot|v, G_\theta)$ is the distribution of $G_\theta$ conditioned on video $v$. $c$ denotes a sampled comment. $V$ is videos from the training dataset.

\begin{table*}[t]
    \centering
    \caption{Statistics of ViCo-20k and its comparison with existing video-language datasets.}
    \begin{tabular}{l l c c c c c} 
    \toprule
    Dataset & Text Type  & \#Videos & \#Texts  & \#Text per Video & Language & w/ likes\\
    \midrule
    MSRVTT~\cite{xu2016msrvtt} & caption & 10k & 200k & 20.0 & English & No\\
    DiDeMo~\cite{anne2017didemo} & caption & 27k & 41k  & 1.5 & English & No\\ 
    ActivityNet Captions~\cite{krishna2017actnetcaption} & caption &100k & 100k & 1.0 & English & No\\
    \midrule
    LiveBot~\cite{ma2019livebot} & video barrage & 2.3k & 89.5M & 39k & Chinese & No\\
    VideoIC~\cite{wang2020videoic} & video barrage & 4.9k & 533M & 108k & Chinese & No \\
    VTC~\cite{hanu2022vtc} & comment & 339k & 4.7M & 13.8 & English & No \\
    \midrule
    ViCo-20k & comment & 20k & 3M & 150 & English & Yes\\
    \bottomrule
    \end{tabular}
    \label{tab:dataset}
\end{table*}

\begin{figure}
    \centering
    \includegraphics[width=\linewidth]{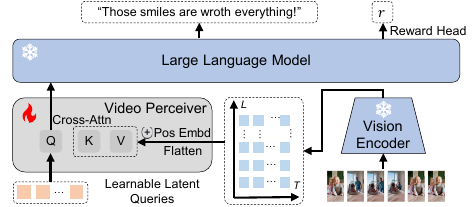}
    \caption{The model architecture of ViCo, which builds on a large language model (LLM) and a large vision model (LVM), with a Video Perceiver module to extract visual information from spatiotemporal patch features and map them to the embedding space of LLM. When used as a generator, it predicts words autoregressively conditioned on a video; and when used as a reward model, it outputs a scalar value for an input video-comment pair.}
    \label{figs:model}
\end{figure}

\subsection{Model Architecture}
We use a video perceiver module to extract useful visual information from the spatiotemporal patch features extracted by an LVM,  and then map visual information to the embedding space of an LLM.
Such an architecture can leverage the powerful representation extracted by LVM and the reasoning and generation ability of LLM. Different from previous LLM based multimodal generators,
our model utilizes a generator-reward-reinforced learning method and processes videos and comments in two modes. When used as a generator, it generates comments conditioned on a given video autoregressively; when used as a reward model, it outputs a scalar reward for a video-comment pair.

\textbf{Video Perceiver Module.}
To bridge the semantic gap between LLM and LVM, we design a video perceiver module inspired by Flamingo~\cite{alayrac2022flamingo} to extract a fixed number of embeddings from a video to represent the most useful information for comment generation and map them to the embedding space of LLM.
Given a video, we first uniformly sample $T$ frames, and use LVM to extract features for each frame to obtain $L$ patch features.
Then we add a spatiotemporal positional embedding to each patch and flatten the patch features of each frame to a sequence of $T\times L$ embeddings.
As there are typically thousands of tokens to represent a video, mapping all of them to the LLM will cause a huge computational burden. The idea of the video perceiver module is to initialize some learned tokens to serve as queries, and the spatiotemporal patch features are used as keys and values. 
During the training process, these learned tokens learn to extract the most useful information from the redundant spatiotemporal patch features and map them to the embedding space of LLM.

\textbf{Formulation of Comment Generator.}
For the generator $G_\theta$, we use it to generate comments conditioned on the video inputs.
After extracting features from the video by LVM and the video perceiver module, these features are projected to the embedding space of the LLM. These tokens act as soft prompts for LLM to generate comments conditioned on video information.

\textbf{Formulation of Reward Model.}
We use a reward model to output a scalar value to evaluate the engagement of video comments.  
For the reward model $R_\theta$, we remove the final embedding-to-token layer of LLM and use a linear head to project the final embeddings to a scalar value. $R_\theta$ process video the same as $G_\theta$, but with an additional comment as input, and then outputs a scalar value for them.

\section{Experiments}

\begin{table*}[t]
\begin{center}
\caption{Results of objective evaluation of generated video comments.}
\resizebox{\linewidth}{!}{
\begin{tabular}{lcccccccc}
\hline
\multicolumn{1}{c}{\multirow{2}{*}{Method}} & \multicolumn{2}{c}{Relevance} & & \multicolumn{2}{c}{Diversity} & & Engagement &\\
\cmidrule{2-3}
\cmidrule{5-6}
\cmidrule{8-9}
 & BLEU $\uparrow$ & ROUGE $\uparrow$ & & num. Bigrams $\uparrow$ & self-CIDEr $\uparrow$ & & AvgReward $\uparrow$ &\\ 
\hline
UT-ResNet~\cite{ma2019livebot}    &12.08 &11.46 & &70k &79.75 & &0.12\\
UT-CLIP~\cite{ma2019livebot}      &14.29 &11.97 & &72k &79.46 & &0.12 \\
CLIP4Caption~\cite{tang2021clip4caption} &14.27 &12.25 & &71k &79.10 & &0.14 \\
GIT-2~\cite{wang2022git}        &21.13 &14.17 & &84k &80.95 & &0.15 \\
\hline
ViCo-random w/o RL  &21.21 &14.36 & &82k  &80.31 & &0.14 \\
ViCo-unique w/o RL &28.36 &15.78 & &110k &84.81 & &0.16\\
ViCo-full version  &\textbf{47.87} &\textbf{18.50}  &  &\textbf{242k}  &\textbf{92.51} & &\textbf{0.44}\\
\hline
\end{tabular}}
\label{tab:quantitative_result}
\end{center}
\end{table*}

\subsection{Datasets}
To facilitate future research, we collect a large video-comment dataset (ViCo-20k) with 20k videos with 3M comments from a popular video website. 
As shown in Fig.~\ref{figs:task_case}, each comment is accompanied by the number of likes from audiences, which can be used as a proxy to indicate their engagement.
To improve the diversity of the dataset, we collect videos belonging to 15 popular categories (e.g., travel, baby) and balance their distribution. 

In Tab.~\ref{tab:dataset}, we show statistics of ViCo-20k and comparisons with other related datasets. 
We can find that videos are paired with much more comments compared with video captions which shows the diversity and difficulty brought to model training.
LiveBot~\cite{ma2019livebot} and VideoIC~\cite{wang2020videoic} are two similar datasets that focus on less common “video barrage” (i.e. time-synchronous) type of comments, which are widely presented in Chinese video platforms but less common in English ones.
VTC~\cite{hanu2022vtc} is closer to our dataset. However, our dataset provides user likes for the first time. 
These likes reflect the engagement of the comments, which can serve as a proxy for comment engagement and enhance the quality of generated comments using our methods. Moreover, each video in our dataset has more comments than VTC. Since video comments are subjective, having more comments is beneficial for model learning.

\subsection{Experimental Setup}

\noindent\textbf{Generator Training.} We select the videos that have more than 10 English comments from ViCo-20k and obtain 14k videos. For each video, we sample a comment of this video using random sampling or our proposed uniqueness-guided sampling. We uniformly sample 12 frames from a video and use CLIP-L/14~\cite{Alec2021CLIP} to extract patch features for each frame. We set 8 layers for the video perceiver module and 40 learnable queries empirically. We use an AdamW optimizer with a learning rate of 5e-6 and warm up the learning rate for 10 epochs. We train our model with a batch size of 256 and train for 100 epochs.

\noindent\textbf{Details of Uniqueness-Guided Sampling.} We retrieve the top 10 similar videos and use their comments as candidates, then we retrieve the top 20 similar comments and calculate BERT similarity for each comment as the uniqueness metric. To ease the model training, we set a 5-level uniqueness sampling schedule by gradually increasing the uniqueness of sampled comments. Thus the difficulty of model training is gradually increased. We empirically set the initial $a$ to 0.6 and increased it by 0.2 for every 20 epochs.

\noindent\textbf{Reward Model Training.}
During the reward model training, we use temporal debiasing to process ViCo-20k, and finally, we get 8.5k videos with 988k comparison pairs of comments. 
The optimization setting is the same as the generator. We train the reward model with a batch containing 128 comparison pairs and train for 10 epochs.

\noindent\textbf{Reinforcement Learning.} For the reinforcement learning stage, 
we adopt PPO~\cite{Schulman2017PPO} algorithm in this stage as~\cite{openai2022chatgpt,ouyang2022instructgpt}. We initialize the policy from the initial generator and the value function from the engagement reward model for PPO. We normalize the reward model outputs such that the rewards achieve a mean score of 0. We use a batch size of 128, a learning rate of 1.4e-5, and an initial KL coefficient of 1.0. We train the model for 5 epochs.

\subsection{Evaluation}
We compare our approach with comment generation models and video caption models, as well as the  variants of our model as follows:
\begin{itemize}
    \item \textbf{Comment Generation}: ~\cite{ma2019livebot,marrese2022open-commentary} adopts a Unified Transformer (UT) encoder-decoder model for live video comment generation, as our task does not need other comments as input, we omit the text encoder. ~\cite{wang2020videoic} proposes a temporal relation prediction task for live video commenting on the top of UT, but it is specifically designed for live video and cannot be applied to our task. These models use ResNet~\cite{he2016resnet} to extract features, we also include the results of using CLIP feature considering its better representation.

    \item \textbf{Video Caption}: As the previous models~\cite{ma2019livebot,marrese2022open-commentary} for comment generation are caption-style models, we compare with recent video captioning models as baselines for comment generation. CLIP4Caption~\cite{tang2021clip4caption} adopts CLIP and transformer encoder-decoder for video caption with an additional video-text matching pre-training stage. We also compare with GIT-2~\cite{wang2022git} which achieves leading performance on video captioning benchmarks. We finetune it on ViCo-20k to generate comments.

    \item \textbf{Our Model}: To demonstrate the effectiveness of our methods, we train our model (\textbf{ViCo}) in three settings: the generator trained with randomly sampled comments (\textbf{ViCo-random w/o RL}), the generator trained with our proposed uniqueness-guided sampling (\textbf{ViCo-unique w/o RL}), and our final model that further trained on the top of initial generator with the guidance of the reward model (\textbf{ViCo-full version}).
\end{itemize}

We conduct both objective and subjective evaluations for the generated comments. For objective evaluation, we evaluate the quality of comments from three perspectives: relevance, diversity, and engagement. For subjective evaluation, we conduct human evaluation to further demonstrate the superiority of our model.
\begin{itemize}
    \item \textbf{Relevance Metrics} compare the n-gram overlap of generated comments with ground truths. We use \textbf{BLEU}~\cite{papineni2002bleu} and \textbf{ROUGE-L}~\cite{lin2004rouge} which are commonly used for caption evaluation as relevance metrics. 
    We sample comments and ask annotators to manually select 10 high-quality comments for each video as ground-truth comments. 
    \item \textbf{Diversity Metrics} measure the diversity and informativeness of generated comments. We use \textbf{num. Bigrams} and \textbf{self-CIDEr}~\cite{Wang_2019_selfcider} as diversity metrics. num. Bigrams uses the number of unique bigrams among generated comments to measure informativeness. Self-CIDEr uses the CIDEr similarity between generated comments for a video to measure diversity.
    \item \textbf{Engagement Metric} is the average reward scores computed by our reward model since it is trained to align the reward of comments to human perception of engagement and is well-aligned with the judgment of human annotators. Although this is not a perfect measure, we use it as a reference for evaluating the engagement of comments.
    \item \textbf{Human Evaluation} is conducted by asking annotators to judge the quality of two groups of comments generated by two models in an A-B test manner. We randomly select 100 videos and generate comments using our model and three baseline models. 
\end{itemize}

\begin{figure*}[t]
    \centering
    \includegraphics[width=\linewidth]{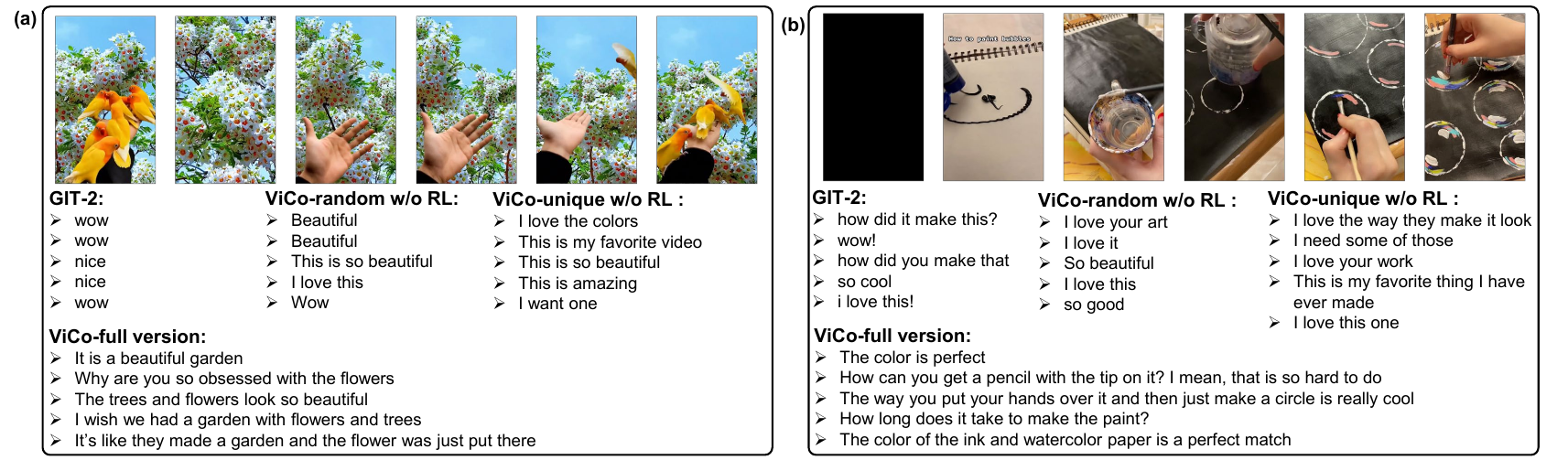}
    \caption{Examples of generated comments by our model (ViCo-full version) and three baselines.}
    \label{figs:qualitive_results}
\end{figure*}

\subsection{Result Analysis}

\noindent\textbf{Effectiveness of the Engagement Reward Model.}
\label{sec:rewardeva}
In order to verify the effectiveness of the reward model, we sample 500 comparison pairs of 100 videos and ask our annotators to select which comment is more engaging, the results show the reward scores are highly aligned with the judgments of our annotators with \textbf{80.4\%} consistency.
The Temporal Debiasing step is essential for training the reward model. Without it, the consistency between the reward model and human judgment is 54.3\%, nearly at a random-chance level.

\noindent\textbf{Results of Objective Evaluation.}
We show the results of the objective evaluation in Tab.~\ref{tab:quantitative_result}. 
Our ViCo-random w/o RL model is almost a captioning model without our designed sampling and reward mechanism. We can also find from the results that it is comparable with GIT-2 in terms of all metrics. UT and CLIP4Caption perform less effectively due to the weaker vanilla Transformer backbone.
With our proposed uniqueness-guided sampling integrated (ViCo-unique w/o RL), the relevance score compared with high-quality comments, and the diversity and engagement scores are all improved. This indicates the effectiveness of our proposed uniqueness-guided sampling for alleviating the phenomenon of the model collapsing to generate only general phrases. 
After adding the reward model for reinforcement learning, our model ViCo-full version improves relevance, diversity, and engagement significantly by enriching the informativeness and engagement of generated comments. 

\noindent\textbf{Examples of Generated Comments.}
In Fig.~\ref{figs:qualitive_results}, we show some examples of generated comments by our model and three baselines. 
In general, models that are trained using random sampling (i.e., GIT-2 and ViCo-random w/o RL) are collapsed to generate general phrases. Our proposed uniqueness-guided sampling (ViCo-unique w/o RL) enhances the richness of the generated comments. After optimizing the initial generator with our engagement reward model, our final model (ViCo-full version) is able to generate relevant, informative, and more engaging comments.
Although GIT-2~\cite{wang2022git} shows strong performance on video captioning, finetuning it on ViCo-20k still makes it collapse to generate general phrases. For example, it generates ``wow'', and ``nice'' for video (a) in Fig.~\ref{figs:qualitive_results}.
ViCo-random w/o RL only generates general while unengaging phrases, such as ``beautiful'', ``wow'' and ``I love it''.
ViCo-unique w/o RL can generate more informative comments than ViCo-random w/o RL.
Among all models, our ViCo-full version model can generate the most engaging comments, focusing more on the unique information in the video. For example, in video (b), our model is able to know that the way the painter draws circles in the video is really cool, which could better engage audiences.

\begin{table}
\centering
\caption{Results of human evaluations. We report the win percentage of ViCo-full version v.s. three baselines. Results are statistically significant with $p< 0.001$.}
\begin{tabular}{cc}
\hline
Model & Win \\
\hline
ViCo-full version v.s. GIT-2 & 84.0\%  \\
ViCo-full version v.s. ViCo-random w/o RL & 83.0\%  \\
ViCo-full version v.s. ViCo-unique w/o RL & 81.7\%  \\
\hline
\end{tabular}
\label{tab:winpercentage}
\end{table}

\noindent\textbf{Results of Human Evaluation.}
In Tab.~\ref{tab:winpercentage}, we show the result of human evaluation.  We compare our ViCo-full version model with three baselines and report the ``win percentage'' for each model pair. We can see that the comments generated by our model are more preferred by people most of the time, with more than 80\%. Compared to GIT-2 and ViCo-random w/o RL, ViCo-unique w/o RL performs better, which is consistent with the objective scores explained above.

\section{Conclusion}

In this paper, we tackle the challenging task of generating engaging video comments with three novel designs. First, we use the number of likes for the comments from online videos as a proxy for human preference on engagement after a debiasing process.
Second, we train a reward model using the above proxy as real-user feedback to automatically evaluate the engagement of comments.
Third, to alleviate the scarcity of engaging video comments, we first train an initial generator on noise data equipped with a uniqueness-guided sampling method, then adopt the reward model to further optimize the generator.
Moreover, we collect a large-scale video-comment dataset (ViCo-20k) to support future studies of video commenting.
Extensive experiments are conducted to verify the effectiveness of each component in our approach.
The results show that our model can generate more informative and engaging comments than baselines.
In the future, we can utilize more powerful LLM further enhance performance, and explore the application of engaging comment generation in scenarios such as multimodal dialogues.

\bibliography{main}

\begin{thebibliography}{32}
\providecommand{\natexlab}[1]{#1}

\bibitem[{Alayrac et~al.(2022)Alayrac, Donahue, Luc, Miech, Barr, Hasson, Lenc,
  Mensch, Millican, Reynolds et~al.}]{alayrac2022flamingo}
Alayrac, J.-B.; Donahue, J.; Luc, P.; Miech, A.; Barr, I.; Hasson, Y.; Lenc,
  K.; Mensch, A.; Millican, K.; Reynolds, M.; et~al. 2022.
\newblock Flamingo: a visual language model for few-shot learning.
\newblock \emph{Advances in Neural Information Processing Systems}, 35:
  23716--23736.

\bibitem[{Anne~Hendricks et~al.(2017)Anne~Hendricks, Wang, Shechtman, Sivic,
  Darrell, and Russell}]{anne2017didemo}
Anne~Hendricks, L.; Wang, O.; Shechtman, E.; Sivic, J.; Darrell, T.; and
  Russell, B. 2017.
\newblock Localizing moments in video with natural language.
\newblock In \emph{Proceedings of the IEEE international conference on computer
  vision}, 5803--5812.

\bibitem[{Brown et~al.(2020)Brown, Mann, Ryder, Subbiah, Kaplan, Dhariwal,
  Neelakantan, Shyam, Sastry, Askell et~al.}]{brown2020gpt3}
Brown, T.; Mann, B.; Ryder, N.; Subbiah, M.; Kaplan, J.~D.; Dhariwal, P.;
  Neelakantan, A.; Shyam, P.; Sastry, G.; Askell, A.; et~al. 2020.
\newblock Language models are few-shot learners.
\newblock \emph{Advances in neural information processing systems}, 33:
  1877--1901.

\bibitem[{Devlin et~al.(2019)Devlin, Chang, Lee, and
  Toutanova}]{Devlin2018bert}
Devlin, J.; Chang, M.-W.; Lee, K.; and Toutanova, K. 2019.
\newblock {\text{BERT}: Pre-training of Deep Bidirectional Transformers for
  Language Understanding}.
\newblock In \emph{Proceedings of the 2019 Conference of the North American
  Chapter of the Association for Computational Linguistics: Human Language
  Technologies, Volume 1 (Long and Short Papers)}, 4171--4186.

\bibitem[{Hanu et~al.(2022)Hanu, Thewlis, Asano, and Rupprecht}]{hanu2022vtc}
Hanu, L.; Thewlis, J.; Asano, Y.~M.; and Rupprecht, C. 2022.
\newblock VTC: Improving Video-Text Retrieval with User Comments.
\newblock In \emph{European Conference on Computer Vision}, 616--633. Springer.

\bibitem[{He et~al.(2016)He, Zhang, Ren, and Sun}]{he2016resnet}
He, K.; Zhang, X.; Ren, S.; and Sun, J. 2016.
\newblock Deep residual learning for image recognition.
\newblock In \emph{Proceedings of the IEEE conference on computer vision and
  pattern recognition}, 770--778.

\bibitem[{Krishna et~al.(2017)Krishna, Hata, Ren, Fei-Fei, and
  Carlos~Niebles}]{krishna2017actnetcaption}
Krishna, R.; Hata, K.; Ren, F.; Fei-Fei, L.; and Carlos~Niebles, J. 2017.
\newblock Dense-captioning events in videos.
\newblock In \emph{IEEE International Conference on Computer Vision (ICCV)},
  706--715.

\bibitem[{Li et~al.(2023)Li, Li, Savarese, and Hoi}]{li2023blip-2}
Li, J.; Li, D.; Savarese, S.; and Hoi, S. 2023.
\newblock Blip-2: Bootstrapping language-image pre-training with frozen image
  encoders and large language models.
\newblock In \emph{International Conference on Machine Learning}.

\bibitem[{Lin(2004)}]{lin2004rouge}
Lin, C.-Y. 2004.
\newblock Rouge: A package for automatic evaluation of summaries.
\newblock In \emph{Text summarization branches out}, 74--81.

\bibitem[{Lin et~al.(2022)Lin, Li, Lin, Ahmed, Gan, Liu, Lu, and
  Wang}]{lin2022swinbert}
Lin, K.; Li, L.; Lin, C.-C.; Ahmed, F.; Gan, Z.; Liu, Z.; Lu, Y.; and Wang, L.
  2022.
\newblock Swinbert: End-to-end transformers with sparse attention for video
  captioning.
\newblock In \emph{Proceedings of the IEEE/CVF Conference on Computer Vision
  and Pattern Recognition}, 17949--17958.

\bibitem[{Ma et~al.(2019)Ma, Cui, Dai, Wei, and Sun}]{ma2019livebot}
Ma, S.; Cui, L.; Dai, D.; Wei, F.; and Sun, X. 2019.
\newblock Livebot: Generating live video comments based on visual and textual
  contexts.
\newblock In \emph{Proceedings of the AAAI Conference on Artificial
  Intelligence}, volume~33, 6810--6817.

\bibitem[{Marrese-Taylor et~al.(2022)Marrese-Taylor, Hamazono, Ishigaki,
  Topi{\'c}, Miyao, Kobayashi, and Takamura}]{marrese2022open-commentary}
Marrese-Taylor, E.; Hamazono, Y.; Ishigaki, T.; Topi{\'c}, G.; Miyao, Y.;
  Kobayashi, I.; and Takamura, H. 2022.
\newblock Open-domain Video Commentary Generation.
\newblock In \emph{Proceedings of the 2022 Conference on Empirical Methods in
  Natural Language Processing}, 7326--7339.

\bibitem[{Mokady, Hertz, and Bermano(2021)}]{mokady2021clipcap}
Mokady, R.; Hertz, A.; and Bermano, A.~H. 2021.
\newblock Clipcap: Clip prefix for image captioning.
\newblock \emph{arXiv preprint arXiv:2111.09734}.

\bibitem[{OpenAI(2022)}]{openai2022chatgpt}
OpenAI. 2022.
\newblock Introducing ChatGPT.

\bibitem[{OpenAI(2023)}]{OpenAI2023GPT4TR}
OpenAI. 2023.
\newblock GPT-4 Technical Report.
\newblock \emph{ArXiv}, abs/2303.08774.

\bibitem[{Ouyang et~al.(2022)Ouyang, Wu, Jiang, Almeida, Wainwright, Mishkin,
  Zhang, Agarwal, Slama, Ray et~al.}]{ouyang2022instructgpt}
Ouyang, L.; Wu, J.; Jiang, X.; Almeida, D.; Wainwright, C.; Mishkin, P.; Zhang,
  C.; Agarwal, S.; Slama, K.; Ray, A.; et~al. 2022.
\newblock Training language models to follow instructions with human feedback.
\newblock \emph{Advances in Neural Information Processing Systems}, 35:
  27730--27744.

\bibitem[{Papineni et~al.(2002)Papineni, Roukos, Ward, and
  Zhu}]{papineni2002bleu}
Papineni, K.; Roukos, S.; Ward, T.; and Zhu, W.-J. 2002.
\newblock Bleu: a method for automatic evaluation of machine translation.
\newblock In \emph{Proceedings of the 40th annual meeting of the Association
  for Computational Linguistics}, 311--318.

\bibitem[{Radford et~al.(2021)Radford, Kim, Hallacy, Ramesh, Goh, Agarwal,
  Sastry, Askell, Mishkin, Clark, Krueger, and Sutskever}]{Alec2021CLIP}
Radford, A.; Kim, J.~W.; Hallacy, C.; Ramesh, A.; Goh, G.; Agarwal, S.; Sastry,
  G.; Askell, A.; Mishkin, P.; Clark, J.; Krueger, G.; and Sutskever, I. 2021.
\newblock Learning Transferable Visual Models From Natural Language
  Supervision.
\newblock In \emph{International Conference on Machine Learning}, 8748--8763.
  {PMLR}.

\bibitem[{Radford et~al.(2019)Radford, Wu, Child, Luan, Amodei, and
  Sutskever}]{radford2019gpt2}
Radford, A.; Wu, J.; Child, R.; Luan, D.; Amodei, D.; and Sutskever, I. 2019.
\newblock Language Models are Unsupervised Multitask Learners.

\bibitem[{Ranzato et~al.(2015)Ranzato, Chopra, Auli, and
  Zaremba}]{ranzato2015sequence}
Ranzato, M.; Chopra, S.; Auli, M.; and Zaremba, W. 2015.
\newblock Sequence level training with recurrent neural networks.
\newblock \emph{arXiv preprint arXiv:1511.06732}.

\bibitem[{Rennie et~al.(2017)Rennie, Marcheret, Mroueh, Ross, and
  Goel}]{rennie2017scst}
Rennie, S.~J.; Marcheret, E.; Mroueh, Y.; Ross, J.; and Goel, V. 2017.
\newblock Self-critical sequence training for image captioning.
\newblock In \emph{Proceedings of the IEEE conference on computer vision and
  pattern recognition}, 7008--7024.

\bibitem[{Schulman et~al.(2017)Schulman, Wolski, Dhariwal, Radford, and
  Klimov}]{Schulman2017PPO}
Schulman, J.; Wolski, F.; Dhariwal, P.; Radford, A.; and Klimov, O. 2017.
\newblock Proximal Policy Optimization Algorithms.
\newblock \emph{ArXiv}, abs/1707.06347.

\bibitem[{Seo et~al.(2022)Seo, Nagrani, Arnab, and Schmid}]{seo2022endcaption}
Seo, P.~H.; Nagrani, A.; Arnab, A.; and Schmid, C. 2022.
\newblock End-to-end generative pretraining for multimodal video captioning.
\newblock In \emph{Proceedings of the IEEE/CVF Conference on Computer Vision
  and Pattern Recognition}, 17959--17968.

\bibitem[{Shuster et~al.(2019)Shuster, Humeau, Hu, Bordes, and
  Weston}]{shuster2019engaging}
Shuster, K.; Humeau, S.; Hu, H.; Bordes, A.; and Weston, J. 2019.
\newblock Engaging image captioning via personality.
\newblock In \emph{Proceedings of the IEEE/CVF Conference on Computer Vision
  and Pattern Recognition}, 12516--12526.

\bibitem[{Tang et~al.(2021)Tang, Wang, Liu, Rao, Li, and
  Li}]{tang2021clip4caption}
Tang, M.; Wang, Z.; Liu, Z.; Rao, F.; Li, D.; and Li, X. 2021.
\newblock Clip4caption: Clip for video caption.
\newblock In \emph{Proceedings of the 29th ACM International Conference on
  Multimedia}, 4858--4862.

\bibitem[{Tsimpoukelli et~al.(2021)Tsimpoukelli, Menick, Cabi, Eslami, Vinyals,
  and Hill}]{tsimpoukelli2021frozen}
Tsimpoukelli, M.; Menick, J.~L.; Cabi, S.; Eslami, S.; Vinyals, O.; and Hill,
  F. 2021.
\newblock Multimodal few-shot learning with frozen language models.
\newblock \emph{Advances in Neural Information Processing Systems}, 34:
  200--212.

\bibitem[{Vedantam, Lawrence~Zitnick, and Parikh(2015)}]{vedantam2015cider}
Vedantam, R.; Lawrence~Zitnick, C.; and Parikh, D. 2015.
\newblock Cider: Consensus-based image description evaluation.
\newblock In \emph{Proceedings of the IEEE conference on computer vision and
  pattern recognition}, 4566--4575.

\bibitem[{Wang et~al.(2022)Wang, Yang, Hu, Li, Lin, Gan, Liu, Liu, and
  Wang}]{wang2022git}
Wang, J.; Yang, Z.; Hu, X.; Li, L.; Lin, K.; Gan, Z.; Liu, Z.; Liu, C.; and
  Wang, L. 2022.
\newblock Git: A generative image-to-text transformer for vision and language.
\newblock \emph{arXiv preprint arXiv:2205.14100}.

\bibitem[{Wang and Chan(2019)}]{Wang_2019_selfcider}
Wang, Q.; and Chan, A.~B. 2019.
\newblock Describing Like Humans: On Diversity in Image Captioning.
\newblock In \emph{Proceedings of the IEEE/CVF Conference on Computer Vision
  and Pattern Recognition (CVPR)}.

\bibitem[{Wang, Chen, and Jin(2020)}]{wang2020videoic}
Wang, W.; Chen, J.; and Jin, Q. 2020.
\newblock Videoic: A video interactive comments dataset and multimodal
  multitask learning for comments generation.
\newblock In \emph{Proceedings of the 28th ACM International Conference on
  Multimedia}, 2599--2607.

\bibitem[{Xu et~al.(2016)Xu, Mei, Yao, and Rui}]{xu2016msrvtt}
Xu, J.; Mei, T.; Yao, T.; and Rui, Y. 2016.
\newblock Msr-vtt: A large video description dataset for bridging video and
  language.
\newblock In \emph{IEEE International Conference on Computer Vision and Pattern
  Recognition (CVPR)}, 5288--5296.

\bibitem[{Zeng et~al.(2021)Zeng, Gao, Xue, and Tu}]{zeng2021plvcg}
Zeng, Z.; Gao, N.; Xue, C.; and Tu, C. 2021.
\newblock PLVCG: A Pretraining Based Model for Live Video Comment Generation.
\newblock In \emph{Pacific-Asia Conference on Knowledge Discovery and Data
  Mining}, 690--702. Springer.

\end{thebibliography}

\end{document}